**Meta Learning for Multi-View Visuomotor Systems**
Benji Alwis

**Abstract**

This paper introduces a new approach for quickly adapting a multi-view visuomotor system for robots to varying camera configurations from the baseline setup. It utilises meta-learning to fine-tune the perceptual network while keeping the policy network fixed. Experimental results demonstrate a significant reduction in the number of new training episodes needed to attain baseline performance.

**Introduction**

Inspired by how humans learn motor skills through trial and error, reinforcement learning is used in end-to-end visuomotor systems [1,2,3] to help robots master complex manipulation tasks based on raw sensory inputs, including visual observations. Online reinforcement learning is deemed impractical because robots need continuous interaction with the environment. Instead, offline learning from pre-recorded demonstrations is a more suitable approach, with imitation learning [4] being a popular choice.

Imitation learning systems work by learning action policies by observing and mimicking the behaviors demonstrated by human or expert agents. Visual observations can originate from multiple cameras, some fixed to the robot while others are external. eye-in-hand robot setups, where a camera is attached to the end-effector, have been proven to be better at generalisation against changes to its workspace such as height or texture changes [5]. However, the limited viewpoint of eye-in-hand setups can be augmented by combining them with external cameras. Unlike robot-mounted cameras, there is less control over the positioning of external cameras, which may change when the deployment environment changes. Retraining from scratch is not practical in these situations since a large number of new demonstrations are required. Recent work has highlighted the limitations caused by the widely used stationary environment policies where policies are learned for static environments [6]. On the other hand, humans can adapt to new situations with relatively less effort. In this paper, our focus is on developing a new approach for rapid adaptation to different camera configurations for robot pick and place tasks.

The approach outlined in this paper uses meta-learning to improve the adaptation of an end-to-end visuomotor system that utilises a convolutional neural network for visual processing and employs imitation learning to acquire action policies. The primary contribution of this paper lies in the incorporation of meta-learning to enable rapid adaptation of end-to-end visuomotor systems to varying camera configurations for robot pick and place tasks. The rest of the paper is organised as follows. We commence with a concise overview of the primary techniques employed, followed by an elaboration of our approach. Subsequently, we present the experimental results and draw our primary conclusions.

**Preliminaries**

Reinforcement Learning

Neural network models used inside end-to-end architectures in robotics map sensory inputs to motor output. In such situations, reinforcement learning based on Partially Observable Markov Decision Processes (POMDPs) can be used to address the inherent uncertainty when robots cannot directly perceive the world's underlying state and it is captured using visual sensors.. Observations produced by the visual sensors are estimates derived from sensor data. When the robot's sensor information is incomplete or noisy, it becomes challenging to ascertain the true state of the environment. Unlike standard Markov Decision Processes (MDPs), which assume full observability, POMDPs take into account the partial observability of the environment. They enable robots to make optimal decisions by considering the uncertainty in their observations and selecting actions that maximise their expected cumulative reward over time.

## Meta Learning

Traditional deep learning faced a significant limitation: once a neural network was trained for a specific task, it could not be readily repurposed for closely related tasks without undergoing complete retraining from scratch. This issue is precisely what meta-learning aims to address. It does so by making use of a shared set of parameters among a group of interconnected tasks, enabling the model's reuse for new, related tasks.

Meta-learning algorithms maintain two sets of parameters: one set is specific to individual tasks (inner parameters), while the other set is universal across all tasks, termed "meta-parameters" or outer parameters, which are determined through learning across a range of related tasks.

In this paper, we employ the Model-Agnostic Meta-Learning (MAML) algorithm [7], which can be summarised as a four-step process.

- **Initialisation**: Start with random model parameters.
- **Meta-Training**: In each iteration, sample a task, split its dataset into support and query sets, compute loss on support for parameter fine-tuning, and update model parameters through gradient descent. Repeat for multiple tasks to update the model across diverse tasks.
- **Meta-Testing**: Post meta-training, the model adapts to new tasks using a small support set, with evaluation on the query set for performance assessment.
- **Iterate**: Repeat the process to improve adaptability.

**Our Approach**

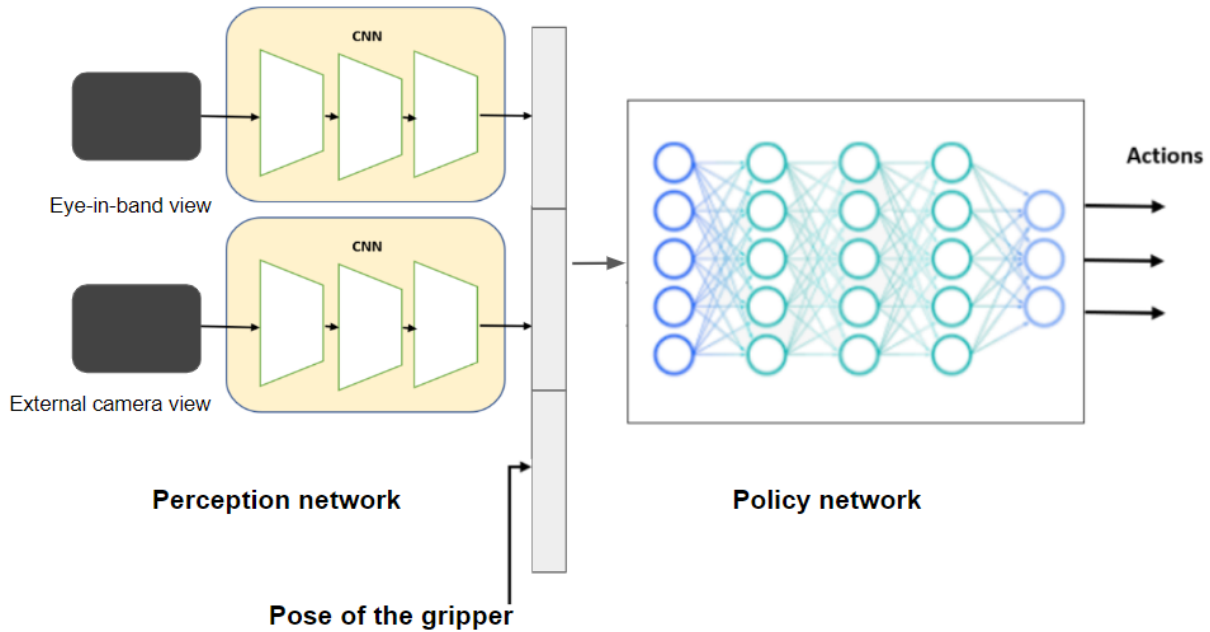

Figure 1 - Overview of our end-to-end visuomotor system.

Our end-to-end visuomotor system, as illustrated in Figure 1, follows, with some customisations, the architecture initially proposed in [8], which has since been adopted by numerous researchers [1,2,3,4]. The primary function of this architecture is to convert images into a distribution over actions. Initially, the images captured by each camera undergo an 18-layer ResNet architecture. To yield a compact feature representation, we have excluded the last fully connected layer typically used for classification tasks. These feature vectors are then concatenated with another vector representing the geometric coordinates of the end-effector. Subsequently, they pass through a neural network implementing the Dagger (Dataset Aggregation) algorithm [9], a technique employed in imitation learning. DAgger involves the agent initially mimicking the expert's actions based on a dataset acquired during the initial training phase. However, as the agent's actions may deviate from the expert's, this can lead to accumulating errors. To mitigate this, DAgger refines the agent's policy iteratively. In each iteration called a DAgger round, the agent collects new data by interacting with the environment and combines it with the previous datasets. This combined data, along with the expert's actions, forms a more comprehensive training dataset. The agent's policy is updated using this aggregated dataset, progressively improving its ability to imitate the expert's behavior. DAgger repeats this process through multiple iterations, enhancing the agent's performance and narrowing the gap between the agent's actions and the expert's, offering a baseline that can be adapted when the external camera configuration changes.

However, when the orientation or position of the external camera is altered, the corresponding ResNet layers produce different feature vectors, even when the state remains unchanged (i.e., there is no change in the arrangement of objects relative to the robot's position and orientation). When the concatenated vector is presented to the DAgger implementation, it may provide incorrect action recommendations. Retraining the network from scratch is impractical. Instead, our approach is to adapt the perception network while keeping the policy network frozen, conceptually similar to [10]. Adaptation occurs within a meta-learning framework. During each meta-learning cycle, the external camera configuration remains fixed, and the same baseline trajectories are followed, with images recorded. The loss function is based on the difference between the new feature vector and the baseline feature vector produced by the perception network (i.e., the difference between the latent states). The training objective is to minimise this difference between the latent states, and training phases are repeated for different camera configurations.

**Experiments**

The main aim of the experiments was to evaluate the effectiveness of utilising meta-learning for quick adaptation across various camera viewpoints. The experiments presented in this paper were conducted using an agent developed with PyBullet, an accessible Python interface to the Bullet physics engine, within an OpenAI gym environment. The agent, specified through a set of URDF files, simulates the Franka Emika Panda robot with a 4-DoF action space, encompassing the position of the end effector in 3D space and a boolean value indicating the gripper state (open or closed). The specified number of episodes generated by the robot simulator was saved in a pickle-formatted file for training the agent.

The offline imitation learning process commenced with the acquisition of a policy through example episodes offered by the Franka Emika Panda robot simulator. This policy learning was achieved via supervised learning, focusing on observation-action pairs. To ensure that the samples are treated as independent and identically distributed (i.i.d), the observation-action pairs were randomly shuffled, mitigating biases between consecutive samples.

The Dagger algorithm was applied with a user-defined number of Dagger rounds(6) and epochs(15) for each round. This approach aimed to progressively enhance policy learning by increasing the number of trajectories through the inclusion of those generated by the agent during interactions with the environment.

The previously outlined steps involve training an agent policy to determine the optimal action for a given observation. However, all observations originated from a single eye-in-hand camera and one external camera with a fixed position and orientation. Our objective was to evaluate the impact of meta-learning for swift adaptation to varying camera configurations.

Once a baseline policy had been trained, the weights in the perception network were stored. Subsequently, observations were recorded for a range of distinct external camera configurations, while keeping the object configurations constant. In total, we employed 20 different object configurations and 6 camera configurations for this purpose. Meta-learning was then conducted using these images. In each round, one camera configuration was set aside for the purpose of adaptation (meta-testing).

In our experiment, we computed the success rate based on the accurate prediction of actions for a given set of observations. As the number of training episodes (sample complexity) increased, the success rate also increased, owing to the expanded size of the training dataset. However, this enhancement came at the cost of having to generate a larger number of training demonstrations, which may be impractical in real-world data collections. Additionally, with the increase in training episodes, the risk of overfitting to the training distribution could also rise. The diagram in Figure 2 illustrates how the success rate varied concerning the number of episodes, both with and without the application of meta-learning. Each data point represents the average success rate from five experimental runs, with the number of training epochs, batch size, and the number of DAgger rounds held constant at 15, 250 and 6 respectively. They show that with meta learning, it was able to reach the success rate of 0.82 with just 40 training episodes. However, it took around 80 episodes to reach similar levels of performance. With over 100 training episodes, it was able to slightly exceed the performance obtained by incorporating meta learning.

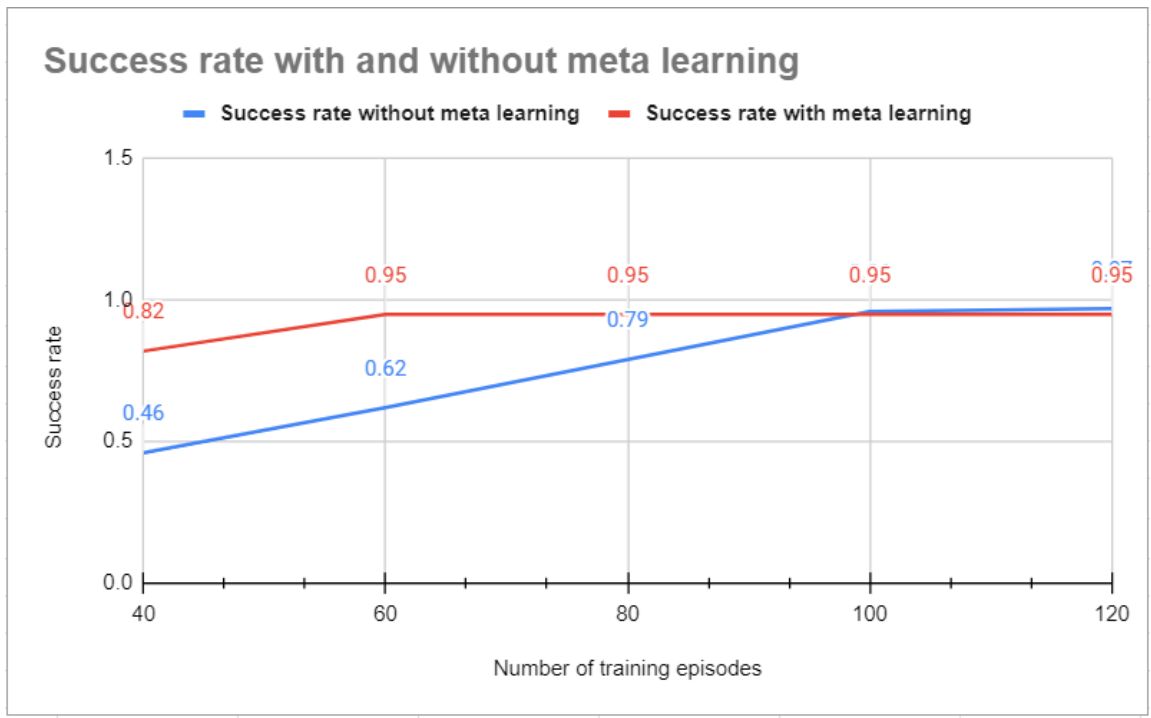

Figure 2 - Experimental results: Success rate vs. Number of training episodes.

**Conclusions**

The objective of this investigation was to evaluate the efficacy of employing meta-learning for quick adaptation across different camera viewpoints. The results indicate that the meta-learning approach is capable of adapting a pre-trained end-to-end visuomotor system with significantly fewer new training episodes in comparison to starting over from scratch. However, with a sufficient number of training episodes, the clean retraining approach was able to match and slightly surpass the performance of the meta-learning approach. This outcome is unsurprising, as a thorough re-training with an adequate volume of training samples suffices, assuming practical feasibility, which is often not the case in many real-world robotics applications.

**Related Work**

End-to-end visuomotor systems have been researched over the past decade [1, 2, 3, 4, 5]. However, our work distinguishes itself by employing meta-learning to address changes in camera configurations within a system that utilises both an eye-in-hand camera and an external camera in tandem. While previous research has explored the use of meta-learning to establish a consistent latent space for adapting end-to-end systems to various visual domains [10] and mobility applications [11], our approach differs in its application of meta-learning to facilitate the adaptation of an end-to-end system to cope with alterations in camera configurations in the context of robot pick and place tasks.